\documentclass{article}
\usepackage{stywhispers,amsmath,epsfig,graphicx}
\usepackage{ctable}
\usepackage{algorithmic, algorithm}
\usepackage{amsmath,amssymb,graphicx,verbatim,enumerate}

\title{Optimal Sparse Kernel Learning for Hyperspectral Anomaly Detection}
\twoauthors
  {Prudhvi Gurram, Heesung Kwon
  }
	{U.S. Army Research Laboratory\\
     ATTN: RDRL-SES-E\\
	2800 Powder Mill RD\\
    Adelphi, Maryland 20783}
  {Zhimin Peng, Wotao Yin 
}
	{Rice University\\
	Department of Computational \\
and Applied Mathematics\\
	Houston, Texas 77005}
\begin{document}
%
\maketitle
\begin{abstract}
In this paper, a novel framework of sparse kernel learning for Support Vector Data Description (SVDD) based anomaly detection is presented. In this work, optimal sparse feature selection for anomaly detection is first modeled as a Mixed Integer Programming (MIP) problem. Due to the prohibitively high computational complexity of the MIP, it is relaxed into a Quadratically Constrained Linear Programming (QCLP) problem.  The QCLP problem can then be practically solved by using an iterative optimization method, in which multiple subsets of features are iteratively found as opposed to a single subset. The QCLP-based iterative optimization problem is solved in a finite space called the \emph{Empirical Kernel Feature Space} (EKFS) instead of in the input space or \emph{Reproducing Kernel Hilbert Space} (RKHS). This is possible because of the fact that the geometrical properties of the EKFS and the corresponding RKHS remain the same. Now, an explicit nonlinear exploitation of the data in a finite EKFS is achievable, which results in optimal feature ranking. Experimental results based on a hyperspectral image show that the proposed method can provide improved performance over the current state-of-the-art techniques.
\end{abstract}
\section{Introduction}
\label{sec:intro}
Feature selection for learning algorithms aims to find a relevant subset of features that can improve the learning performance by discarding features not useful or even harmful for the given tasks. In the case of kernel-based anomaly detection, such as SVDD, the feature selection requires the accurate estimation of the contribution of each feature to the objective function, i.e., the radius of a hypersphere in the RKHS.

In this paper, a new framework of optimal sparse kernel learning for SVDD-based anomaly detection (OSKLAD) is proposed. The proposed OSKLAD optimally extends the feature selection technique used for the kernel-based learning approaches \cite{Tsang} into SVDD-based anomaly detection by fully optimizing the feature selection method for nonlinear kernels in a newly defined finite space called the EKFS \cite{learnKernel}. Hence, the OSKLAD can be considered as a fully optimized version of the wrapper approach to the SVDD-based anomaly detection with nonlinear kernels. The initial objective of the proposed OSKLAD begins with finding a single subset of original features that can be used to build an optimal hypersphere in the RKHS. This objective can be modeled as a Mixed Integer Programming (MIP) problem. However, the MIP problem is NP-hard, and so the MIP model is relaxed into a Quadratically Constrained Linear Programming (QCLP) problem \cite{qclp} by converting the objective function of the MIP problem into lower bounded quadratic inequality constraints. This QCLP problem is yet intractable due to the prohibitively large number of the inequality constraints. 
To address this issue, a cutting plane method based on the \emph{restricted master problem} coupled with Multiple Kernel Leaning (MKL) \cite{Rako2008} is iteratively used. The goal is to find only a small subset of the inequality constraints that are actively used to define the feasible region of the parameters of the given QCLP problem.

The \emph{active} constraints are effectively identified by finding the \emph{most violating constraints} instead whose half-planes maximally violate the corresponding inequality constraints. Therefore, the task becomes finding multiple subsets of \emph{most violated features} associated with the corresponding \emph{most violating constraints} given the objective function, such as the radius of a hypersphere in the RKHS. However, finding the most violating constraints also becomes a combinatorial problem, if nonlinear kernels, such as Gaussian RBF kernel or high order polynomial kernels, are used, due to the prohibitively large number of possible combinations (subsets) of the original features. To tackle this issue, in the proposed OSKLAD, the most violated features are found in the EKFS. The EKFS is a finite space linearly spanned by basis vectors, which are generated by a map, called the Empirical Kernel Map (EKM). It is shown that the EKHS and the corresponding RKHS constructed by using the same kernel function have the same geometrical property. This means that solutions of any optimization problem obtained from either space are identical. In the proposed OSKLAD, the subsets of the most violated features are optimally found in the EKFS since individual feature ranking in terms of contribution to the radius in the EKFS can be performed optimally based on the property of canonical dot product and the finite dimensionality of the space.



\section{Optimal Sparse Kernel Learning}
In this section, we present an optimal sparse kernel learning for anomaly detection (OSKLAD) using SVDD as a basic building block. Inspired by the feature selection approach for the kernel-based classification \cite{Tsang}, the OSKLAD addresses the problem of the optimal feature selection for the SVDD-based anomaly detection. 
The basic formulation of OSKLAD  is to minimize the radius of the enclosing hypersphere while allowing outliers except that in OSKLAD, only a subset of features is used. So, the model is described as a mixed integer programming problem:
\begin{equation}\label{equ:SFSVDD}
    \begin{aligned}
       \min_{\mathbf{d}} \min_{R, \xi_i, \mathbf{a}}  ~ & R^2 + C \cdot \sum_{i=1}^{N} \xi_i\\
        \text{subject to} ~ & \| \Phi(\tilde{\textbf{x}}_i)  - \mathbf{a} \|^{2} \leq R^2 + \xi_i \\
        & \xi_i \geq 0 \\
        & \tilde{\textbf{x}}_i = \textbf{x}_i \odot \textbf{d}, ~  i =1,2,...,N,
    \end{aligned}
\end{equation}
where $\textbf{d}\in \mathbb{D} = \{\textbf{d}| d_j \in \{0,1\},\sum_{j=1}^M \textbf{d}_j = B, j =1,2,...,M \}$, and $\odot$ represents elementwise product. 
Here $B$ is a threshold that controls the number of features that are selected.
If one assumes that $\mathbf{d}$ is fixed in Eq. \ref{equ:SFSVDD}, it turns into a continuous constrained optimization problem just like a standard SVDD. By applying the Langrange multipliers and KKT conditions to it, we can derive the dual problem (similar to standard SVDD) as:
\begin{equation}\label{equ:DualSSVDD}
    \begin{aligned}
        \min_{\mathbf{d}} \max_{\alpha_{i}} & \sum_{i=1}^{N} \alpha_i k(\tilde{\textbf{x}}_i,\tilde{\textbf{x}}_i) - \sum_{i=1}^{N} \sum_{j=1}^{N} \alpha_i
        \alpha_j k(\tilde{\textbf{x}}_i,\tilde{\textbf{x}}_j) \\
        \text{subject to} & \sum_{i=1}^{N} \alpha_{i} = 1 \\
                   & 0 \leq \alpha_i \leq C \\
                   & \tilde{\textbf{x}}_i = \textbf{x}_i \odot \textbf{d}, ~  i =1,2,...,N.
    \end{aligned}
\end{equation}
However, one should notice that Eq. \ref{equ:DualSSVDD} is still a mixed interger programming (MIP) problem due to the last constraint, which is computationally expensive to solve. In order to solve this problem, it can be converted into a Quadratically Constrained Linear Programming (QCLP). 
We define $S(\alpha, \textbf{d})=\sum_{i=1}^{N}\alpha_i k(\tilde{\textbf{x}}_i,\tilde{\textbf{x}}_i)- \sum_{i,j=1}^{N}\alpha_i\alpha_j k(\tilde{\textbf{x}}_i,\tilde{\textbf{x}}_j)$, and introduce an additional parameter $t$ to obtain the QCLP equivalent of (\ref{equ:DualSSVDD}) as follows:
\begin{equation}\label{equ:QCLP}
    \begin{aligned}
        \max_{\textbf{a}, t}~ & t \\
         \text{subject to}  ~ & \sum_{i=1}^{N} \alpha_i = 1 \\
                            ~ & 0 \leq \alpha_i \leq C \\
                            ~ & t \leq S(\alpha, \mathbf{d}),~ \forall ~ \mathbf{d} \in D.
   \end{aligned}
\end{equation}
Though Eq. \ref{equ:QCLP} is convex, a large number of inequality constraints (last condition in Eq. \ref{equ:QCLP}) makes it impractical to be solved by existing techniques. The number becomes huge if the features reside in a high dimensional space. Note that not all the inequality constraints used in Eq. \ref{equ:QCLP} are actively used in defining the feasible region of the optimization problem. In fact, only a small number of the constraints are useful and directly used to solve the optimization problem. Therefore, an iterative algorithm can be used, in which instead of solving Eq. \ref{equ:QCLP} at once, an intermediate solution pair $(t,\alpha)$ is iteratively updated based on a limited subset of previously found active constraints.  This optimization problem is called the \emph{restricted master problem}, which is closely related to the cutting plane algorithm described in \cite{CutPlane}. The \emph{restricted master problem} consists of two steps \cite{Ye}: 1) $(t,\alpha)$ are optimized based on a previously found restricted subset $\mathcal{I}$ of features, which maximally violates the constraints; and 2) a new vector $\textbf{d}$ of the most violated features is obtained based on newly optimized  $(t,\alpha)$ in step 1 and added to the restricted subset $\mathcal{I} = \mathcal{I} \bigcup \textbf{d} $. These two steps are iterated until convergence \cite{Prudhvi2012OSKLAD}. Finding $\textbf{d}$ of the most violated features is detailed in the next subsection.

The intermediate solution pair $(t,\alpha)$ is now obtained from the following optimization problem
\begin{equation}\label{equ:QCLPRMP}
    \begin{aligned}
        \max_{\textbf{a}, t}~ & t \\
         \text{subject to}  ~ & \sum_{i=1}^{N} \alpha_i = 1, \\
                            ~ & 0 \leq \alpha_i \leq C, \\
                            ~ & t \leq S(\alpha, \textbf{d}^{l}),~ \textbf{d}^{l} \in \mathcal{I}.
   \end{aligned}
\end{equation}
Let $\mu_l \geq 0$ be the dual variable for each constraint in Eq. \ref{equ:QCLPRMP}. The Lagrangian of Eq. \ref{equ:QCLPRMP} can be written as:
\begin{equation}\label{equ:QCLPlg}
	\begin{aligned}
		L(t,\mu) = t +\sum_{l=1}^p \mu_{l} S(\alpha, \textbf{d}^{l}).
	\end{aligned}
\end{equation}
By setting $\frac{\partial L}{\partial t} = 0$, we have $\sum_{l=1}^p \mu_l = 1$. The Lagrangian $L(t,\mu)$, after applying this partial KKT condition, can be rewritten as $L(t,\mu)=\sum_{l=1}^p \mu_{l} S(\alpha, \textbf{d}^{l})$, which transforms (\ref{equ:QCLP}) to the following problem:
\begin{equation}\label{equ:SKAD}
    \begin{aligned}
        \max_{\alpha} \min_{\mu}~ &  \sum_{l=1}^p \mu_{l} S(\alpha, \textbf{d}^{l})\\
         \text{subject to}  ~ & \sum_{i=1}^{N} \alpha_i = 1 \\
                            ~ & 0 \leq \alpha_i \leq C ~ \text{for} ~ i=1,2...,N\\
                            ~ & \sum_{l=1}^{p} \mu_l = 1,  \mu_l \geq 0 ~ \text{for} ~ l=1,2...,p.
   \end{aligned}
\end{equation}
One can observe that can be solved using a two-step iterative process to obtain optimal sparse weights of individual kernels $\mu$ and optimal lagrange multipliers $\alpha^*$ (which define the support vectors or the enclosing hypersphere).

\section{Optimal Feature Selection: Finding Maximally Violating Features}
For updating $\mathbf{d}$, the features that maximally violate the last constraint in Eq. \ref{equ:QCLP} need to be determined. Since the goal of Eq. \ref{equ:QCLP} is to maximize $t$, and it is upper-bounded by $S(\alpha, \textbf{d})$ according to the constraint, the features that maximally violate this constraint will minimize $S(\alpha, \textbf{d})$. One has to solve the following optimization problem:
\begin{equation}\label{equ:updateD}
    \begin{aligned}
        \min_{\textbf{d}} & ~S(\alpha, \textbf{d})\\
        \text{subject to} & ~\sum_{i=1}^M d_i = B\\
                          & ~ d_i \in \{0,1\}.
    \end{aligned}
\end{equation}
In this section, we describe the method to find these feature vectors for both linear kernel and non-linear kernel.

\subsection{Linear Kernel}
If a linear kernel is used, since $k \left(\textbf{x}_i,\textbf{x}_j\right) = \left\langle \textbf{x}_i, \textbf{x}_j \right\rangle$, we have $S(\alpha, \textbf{d})=\sum_{j=1}^{M}d_j c_j$, where $c_j = \sum_{i=1}^N\alpha_i x_{ij}^2 + (\sum_{i=1}^N \alpha_i x_{ij})^2$. $S(\alpha, \textbf{d})$ is a linear function of $\textbf{d}$. Once we have optimal support vectors, the global solution of $\textbf{d}$ can be easily obtained by sorting $c_j$'s in ascending order and setting the first $B$ corresponding elements in $\textbf{d}$, $d_j$ to $1$ and the rest to $0$. Once the optimal feature subset is chosen for a kernel, optimal $\alpha$ and $\mu$ are updated by solving Eq. \ref{equ:SKAD}. These two steps are repeated until the algorithm converges.
%

\subsection{Non-linear Kernel}
If a Gaussian RBF kernel is used, $S(\alpha, \textbf{d})$ is not a linear function of $\textbf{d}$. We cannot solve the problem in Eq. \ref{equ:updateD} optimally because of the large number of combinations of features that have to be considered. So, the data is tranformed from infinte dimensional RKHS into another space called \emph{empirical kernel feature space} (EKFS) with finite dimensionality using \emph{empirical kernel map} (EKM). This will allow us to select subsets of features optimally while still preserving the nonlinear correlations among the features. For a given set of training data points $\{ \textbf{x}_i \}_{i=1}^{n}$, the map defined by
\begin{equation}
\Phi_n : \mathbb{R}^n \rightarrow \mathbb{R}^n \textrm{ } \text{where} \textrm{ } \textbf{x} \mapsto k \left( \cdot,\textbf{x} \right) = \left( k \left( \textbf{x}_1,\textbf{x} \right),\ldots,k \left( \textbf{x}_n,\textbf{x} \right) \right)^T
\label{eq:ekm}
\end{equation}
is called the EKM with respect to $\{\textbf{x}_i\}_{i=1}^{n}$ \cite{learnKernel}. However, the kernel function $k$ used to build kernel matrices in previous subsections cannot be represented using $\Phi_n$, since they do not form an orthonormal system. The dot product to use in the representation of $k$ is the not the canonical dot product in the EKFS $\mathbb{R}^n$. In order to turn $\Phi_n$ into a feature map associated with $k$, EKFS is endowed with a dot product $\left\langle \cdot,\cdot\right\rangle_n$ such that $k(\textbf{x}_i,\textbf{x}_j) = \left\langle \Phi_n \left ( \textbf{x}_i \right ), \Phi_n \left ( \textbf{x}_j \right ) \right\rangle_n$. After analyzing certain conditions using this equality as shown in \cite{learnKernel}, the dot product $\left\langle \cdot,\cdot\right\rangle_n$ can be converted to a canonical dot product by merely whitening the EKFS and using the new basis functions as features. It can be represented as
\begin{equation}
k(\textbf{x}_i,\textbf{x}_j) = \left\langle \Phi_n^w \left ( \textbf{x}_i \right ), \Phi_n^w \left ( \textbf{x}_j \right ) \right\rangle,
\label{eq:dotpro}
\end{equation}
where the feature map in whitened EKFS is given by
\begin{equation}
\Phi_n^w :\textrm{ } \textbf{x} \mapsto K^{-\frac{1}{2}} \left( k \left( \textbf{x}_1,\textbf{x} \right),\ldots,k \left( \textbf{x}_n,\textbf{x} \right) \right)^T.
\label{eq:wekm}
\end{equation}
where $K$ is the Gram matrix and $K_{i,j} = k(\textbf{x}_i,\textbf{x}_j)$.
The kernel function in Eq. \ref{eq:dotpro} is used to build the kernel matrices in Eqs. \ref{equ:DualSSVDD}-\ref{equ:updateD}. Hence, the feature subset selection problem turns exactly into (\ref{equ:updateD}) (linear version) except for the fact that in this case the features are selected in EKFS. Similar to the OSKLAD with a linear kernel, the overall Optimal Sparse Kernel Learning for Anomaly Detection (OSKLAD) in the EKFS is described in Algorithm 1.

\begin{algorithm}
\caption{OSKLAD with nonlinear kernel}
\label{Alg:SSVDDNL}
\begin{algorithmic}[2]
\item[1:] Map the data points into the EKFS by using a certain kernel $k$
\item[2:] Initialized: $\alpha = \frac{1}{N}\mathbf{1}$, find the maximally violating feature subset $\textbf{d}$, and set $\mathcal{I}=\{\textbf{d}\}$.
\item[3:] Run SKAD based on the kernel matrices generated by $\mathcal{I}$ and optimize for $\alpha$ and $\mu$.
\item[4:] Find the next maximally violated feature subset $\mathbf{d}$ based on the current $\alpha$ and $\mu$ and set $\mathcal{I}=\mathcal{I}\bigcup\{d\}$.
\item[5:] Repeat steps 3-4 until convergence.
\end{algorithmic}
\end{algorithm}

\section{Simulation Results}
\label{sec:format}

In this section, the performance of OSKLAD is evaluated on a hyperspectral digital imagery collection experiment(HYDICE) image, which contains 30 small painted pannels located in the background. 
We chose a small patch (69 pixels $\times$ 10 pixels) as the background data set, which is used to obtain the radius $R$ and the center of the hypersphere. The distance of each test pixel in the image to the center of the hypersphere is determined. If the distance is greater than $R$, the pixel is considered as an anomaly, otherwise, it is a background pixel. In our experiments, the performance of SVDD, SKAD \cite{SKAD}\footnote{Sparse kernel-based anomaly detection (SKAD) has been developed by two of the current authors} and OSKLAD with both linear and Gaussian RBF kernels are compared with one another. For SVDD and SKAD, both linear and Gaussian RBF kernel are used in the input space. For OSKLAD with linear kernel, feature selection is performed in the input space. However, for OSKLAD with Gaussian RBF kernel, the input vector is first mapped into EKFS using EKM. At this point, we can just use linear kernel in EKFS, which translates to using Gaussian RBF kernel in the input space as described in the previous sections. The kernel bandwidth parameter is determined by implementing the minimax technique on randomly selected 10 regions of the image to represent the background as done in \cite{banerjee2006support}. The same value is used over all the test pixels in the image for all the algorithms.

\begin{figure}[htb]
\begin{minipage}[b]{.48\linewidth}
  \centering
 \centerline{\epsfig{figure=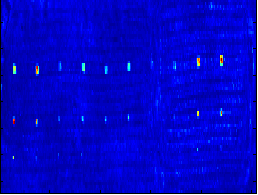,width=3.0cm}}
  \centerline{(a) SVDD -- linear kernel}\medskip
\end{minipage}
\hfill
\begin{minipage}[b]{0.48\linewidth}
  \centering
 \centerline{\epsfig{figure=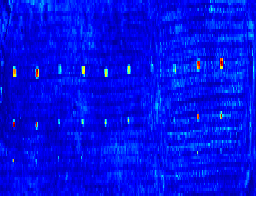,width=3.0cm}}
  \centerline{(b) SVDD -- RBF kernel}\medskip
\end{minipage}
\begin{minipage}[b]{.48\linewidth}
  \centering
 \centerline{\epsfig{figure=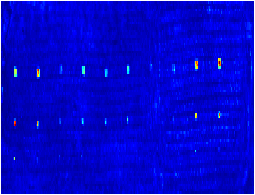,width=3.0cm}}
  \centerline{(c) SKAD -- linear kernel}\medskip
\end{minipage}
\hfill
\begin{minipage}[b]{0.48\linewidth}
  \centering
 \centerline{\epsfig{figure=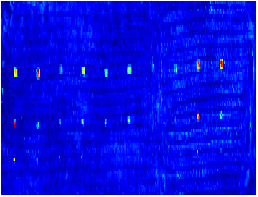,width=3.0cm}}
  \centerline{(d) SKAD -- RBF kernel}\medskip
\end{minipage}
\begin{minipage}[b]{.48\linewidth}
  \centering
 \centerline{\epsfig{figure=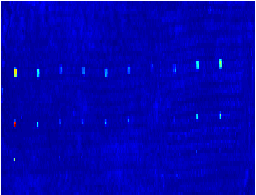,width=3.0cm}}
  \centerline{(e) OSKLAD -- linear kernel}\medskip
\end{minipage}
\hfill
\begin{minipage}[b]{0.48\linewidth}
  \centering
 \centerline{\epsfig{figure=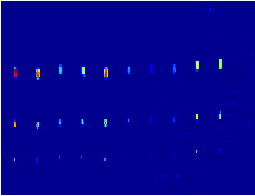,width=3.0cm}}
  \centerline{(f) OSKLAD -- EKFS}\medskip
\end{minipage}
\caption{Anomaly detection results of the HYDICE image using SVDD, SKAD and OSKLAD}
\label{fig:res}
\end{figure}

The number of features used for each hypersphere of SKAD with both linear and Gaussian RBF kernel and OSKLAD with linear kernel is 75, which is half of the total number of features. For OSKLAD in the EKFS, the total number of features available after mapping the pixels from input space to EKFS is reduced to 96, and we have used 48 features for each hypersphere. Fig.\ref{fig:res} shows the anomaly detection results for SVDD, SKAD and OSKLAD with both linear and Gaussian RBF kernels. The value of each pixel in the results is the ratio of the distance between the pixel and the radius of the hypersphere. For comparison, we normalize the scaled in all the resulting images to be between 0 and 1. One can see that all the six methods are able to identify the first two rows of anomalies, but OSKLAD in EKFS can identifies anomalies with much less noise (clean background) and it is also able to detect the small targets in the third row.

\section{Conclusions}
In the proposed work, to achieve optimality in kernel-based feature selection for anomaly detection using SVDD, the QCLP problem is optimally solved in a new finite space called the Empirical Kernel Feature Space (EKFS) instead of the RKHS. Experimental result show that by optimally selecting features, significant improvements can be made in hyperspectral anomaly detection in EKFS rather than the original input space.


\bibliographystyle{IEEEbib}
\bibliography{strings,myref}

\end{document}